\documentclass[10pt, a4paper]{article}

\usepackage{amsmath}
\usepackage{amssymb}
\usepackage{multirow}

\usepackage{enumitem}
\usepackage{booktabs}
\usepackage{tcolorbox}
\usepackage{float}

\usepackage{xurl}
\usepackage[hidelinks]{hyperref}

\usepackage[final]{lrec2026} 

\title{Multilingual Cognitive Impairment Detection in the Era of Foundation Models}

\name{%
\parbox{\textwidth}{\centering\bfseries
Damar Hoogland$^{1}$ \quad
Boshko Koloski$^{1,2}$ \quad
Jaya Caporusso$^{1,2}$ \quad
Tine Kolenik$^{3}$ \\
Ana Zwitter Vitez$^{4}$ \quad
Senja Pollak$^{1}$ \quad
Christina Manouilidou$^{4}$ \quad
Matthew Purver$^{1,5}$
}%
}

\address{%
\parbox{\textwidth}{\centering\small
$^{1}$ Jožef Stefan Institute, Ljubljana, Slovenia\\
$^{2}$ Jožef Stefan International Postgraduate School, Ljubljana, Slovenia\\
$^{3}$ Institute of Synergetics and Psychotherapy Research, Paracelsus Medical University, Salzburg, Austria\\
$^{4}$ University of Ljubljana, Ljubljana, Slovenia \\
$^{5}$ Queen Mary University of London, London, UK
}%
}

\abstract{
We evaluate cognitive impairment (CI) classification from transcripts of speech in English, Slovene, and Korean. We compare zero-shot large language models (LLMs) used as direct classifiers under three input settings—transcript-only, linguistic-features-only, and combined—with supervised tabular approaches trained under a leave-one-out protocol. The tabular models operate on engineered linguistic features, transcript embeddings, and early or late fusion of both modalities. Across languages, zero-shot LLMs provide competitive no-training baselines, but supervised tabular models generally perform better, particularly when engineered linguistic features are included and combined with embeddings. Few-shot experiments focusing on embeddings indicate that the value of limited supervision is language-dependent, with some languages benefiting substantially from additional labelled examples while others remain constrained without richer feature representations. Overall, the results suggest that, in small-data CI detection, structured linguistic signals and simple fusion-based classifiers remain strong and reliable signals.
\\ \newline  }

\begin{document}

\maketitleabstract

\section{Introduction}

Cognitive impairment (CI) refers to a state in which a person's cognitive functioning is below the expected level and is a diagnosable condition \cite{ray2014dementia}. CI can involve varying degrees of deterioration of cognitive abilities such as memory, attention, executive functioning, and language, and it is often associated with neurodegenerative diseases like Alzheimer’s disease (AD) and other conditions that cause dementia \cite{morley2018}. Although relatively advanced CI associated with such diseases is often preceded by a mild cognitive impairment (MCI) phase, not all individuals with MCI progress to dementia \cite{petersen2016mild}. Early identification of CI is essential to enable timely and appropriate clinical intervention, patient support, and participation in preventive or therapeutic programmes \cite{livingston2024}.

Traditional diagnostic assessments for cognitive impairment include neurophysiological tests \cite{nasreddine2005}, clinical and functional assessments \cite{obryant2008}, neuroimaging and biomarker assessments \cite{hampel2018}, and clinical interviews and observation \cite{mckhann2011}. Many of these assessments involve language, as individuals with CI frequently exhibit lexical retrieval difficulties, semantic degradation, syntactic simplification, and reduced discourse organisation, reflecting underlying deterioration in semantic memory and executive control \citep{taler2008, fraser2015, boschi2017}. For example, picture description tasks such as the Cookie Theft task \cite{goodglass1983} prompt individuals to describe a complex visual scene. The resulting descriptions enable qualitative and quantitative assessment of language production, including lexical retrieval, syntactic formulation, fluency, informativeness, and narrative organisation.

However, traditional diagnostic tools can be limited in their ability to detect early cognitive changes \cite{trzepacz2015} and often require in-person administration by trained professionals, making them resource-intensive, time-consuming, and impractical for frequent or large-scale screening \cite{slegers2018}. Their outcomes can furthermore be affected by education and language background, introducing cultural and linguistic bias \cite{ramoshenderson2025}. Finally, because they are typically administered infrequently in clinical settings, they provide only snapshots of cognition rather than a continuous measure of change \cite{patnode2020}.

Computational methods—particularly those leveraging Natural Language Processing (NLP), Machine Learning (ML), and Deep Learning (DL)—address several limitations of traditional assessments by enabling automated and fine-grained analysis of spontaneous speech. Such approaches can capture subtle linguistic and discourse changes that may precede clinical diagnosis, operate remotely and non-invasively, and allow for repeated or continuous monitoring over time \cite{delafuentegarcia2020}. While computational approaches risk inheriting or amplifying linguistic and cultural biases present in existing assessments, they also have the potential to support diagnosis by trained clinicians when developed and evaluated responsibly.

With recent advances in pretrained foundation models and increased accessibility of computational resources, research on automatic CI detection from language has begun shifting from classical ML approaches that rely on expert-selected symbolic features towards methods built around pretrained language models and general-purpose tabular predictors (reviewed in Section~\ref{sec:related-work}). In this study, we compare inference-only foundation-model baselines (multilingual Large Language Models (LLMs) and tabular foundation models) against classical ML and fusion-based approaches across three languages, under unified zero-shot, few-shot, and leave-one-out evaluation protocols.

\section{Related Work}
\label{sec:related-work}

Many studies investigating the detection and prediction of cognitive decline have employed classical ML approaches \cite{huang2024, kaser2024}. For example, \citet{luz2021} employed a range of classical ML models---including linear discriminant analysis, decision trees, $k$-nearest neighbours, random forests, and support vector machines---to classify Alzheimer’s vs.\ healthy speech and to predict scores obtained with the Mini-Mental State Examination test. These models were built on manually engineered acoustic and linguistic feature sets.

More recently, others have moved to using Large Language Models (LLMs) for feature extraction. For example, \citet{dearribaperez2024} automatically extracted a large set of high-level, content-independent linguistic features from free dialogues using ChatGPT\footnote{https://chat.openai.com/} prompts, alongside traditional $n$-gram features. These features were then analysed, selected, and used to train classical ML classifiers to detect cognitive decline. Other studies employed DL techniques, such as BERT-based transformer classifiers \cite{mao2022, ilias2022, pahar2025, zhu2022}.

Large Language Models (LLMs) are increasingly employed as direct classifiers for cognitive decline detection \citep[e.g.,][]{jiang2026llms}. \citet{zheng2024} used pre-trained LLaMA-2\footnote{https://www.llama.com/llama2/} models with prompt engineering, Low-Rank Adaptation (LoRA) fine-tuning, and conditional learning to classify AD from speech transcripts of the ADReSS dataset \cite{luz2020address}. \citet{guan2025} presented CD-Tron, a system built on the clinical LLM GatorTron\footnote{https://huggingface.co/UFNLP/gatortron-base}, fine-tuned on labelled electronic health record note sections to detect early cognitive decline, and reported substantial improvements over smaller transformers and GPT-4. \citet{botelho2024} used LLMs both as predictors and as feature extractors, prompting LLMs to produce interpretable macro-descriptors (e.g., coherence, lexical diversity, word-finding difficulty) which were then used as inputs to simple classifiers for AD detection.

Most studies reviewed above, both classical ML and LLM-based approaches, have focused on monolingual settings, with the majority using only English data. This makes it difficult to assess how well these paradigms transfer across languages and elicitation paradigms.

Tabular foundation models have recently been proposed as general-purpose predictors for structured data, enabling strong performance on small tabular datasets via in-context learning and reducing the need for task-specific training \cite{hollmann2025accurate}. In the cognitive decline domain, \citet{ding2026longitudinalprogressionpredictionalzheimers} apply TabPFN to longitudinal AD modelling on the TADPOLE benchmark, predicting clinical diagnosis and cognitive scores from tabular patient records. Related work explores TabPFN-based approaches for dementia-related prediction in other neurodegenerative settings, such as predicting Parkinson’s disease dementia using a hybrid LightGBM--TabPFN model with SHAP-based interpretability \cite{tran2024predicting}. These studies do not use language data directly, and comparisons to language-based CI detection remain limited.

\subsection{This study}

Prior work has not systematically compared (i) classical ML models with expert-assisted linguistic features, (ii) embedding-based tabular models, (iii) tabular foundation models, and (iv) prompted LLM classifiers under a unified protocol and across multiple languages. In the present study, we conduct within-language experiments for three languages—English, Slovene, and Korean—and compare symbolic-feature-based models, embedding-based models, fusion strategies, and zero-shot LLM baselines under unified leave-one-out, zero-shot, and few-shot protocols. We further analyse representation alignment between symbolic features and embeddings to understand when and why multimodal fusion helps in small-data CI detection.

Our research questions (RQs) are as follows.

\begin{itemize}
    \item {\bf RQ1}: How well do LLMs (specifically, gpt-oss-20b and med-gemma-27b) discriminate CI vs.\ Healthy Control (HC) participants across three languages under zero-shot prompting?

    \item {\bf RQ2}: How sensitive to the input modality (transcript-only, linguistic-features-only, or transcript+features) is the performance of LLMs?

    \item {\bf RQ3}: Do expert-assisted symbolic linguistic features improve CI classification when paired with tabular models (TabPFN, RealMLP, and classical baselines) compared to (i) embedding-only representations and (ii) LLM-based classifiers, and are gains consistent across languages and evaluation paradigms?

    \item {\bf RQ4}: Which integration strategy yields the best and most stable performance across languages among embeddings-only, symbolic-features-only, and multimodal fusion (normalised concatenation / feature reweighting, or late fusion), under both leave-one-out and few-shot evaluation?
\end{itemize}

\section{Data and preprocessing}

\subsection{Datasets}
\label{sec:data}

To evaluate cross-linguistic generalisability and performance stability, we ran parallel experiments on three languages: English, Slovene, and Korean. The English and Slovene datasets were obtained from corpora of recordings of picture description tasks, while the Korean data came from a corpus of structured interviews. The English and Slovene datasets include participants with AD and HCs, while the Korean dataset includes participants with MCI and HCs. In all experiments we treat the positive class as \textsc{Patient} (AD or MCI, depending on the dataset) and the negative class as \textsc{Control} (HC).

The included datasets are listed below, and Table~\ref{tab:participants} summarises the number of participants and diagnostic labels per dataset.

\begin{description}
    \item[English:] The English dataset consists of a subset of Cookie Theft Picture Descriptions from the Pitt Corpus (\citealp{becker1994}; the original corpus is available on DementiaBank, \citealp{macwhinney2011}), pre-processed for the ADReSS challenge \cite{luz2020address}. It includes participants with AD and Healthy Controls (HCs).

    \item[Slovene:] The Slovene dataset was collected as part of the \href{https://sites.google.com/view/coglitreat/home}{CogLiTreat project}, which investigated behavioural and transcranial magnetic stimulation interventions for language disorders. It includes recordings and transcripts of Slovene AD patients and control participants from the Ljubljana region who described the New Cookie Theft picture \cite{berube2019}.
    The participants’ responses were recorded and the detection of speech and silence was performed automatically using Praat \cite{boersma2021}. The recordings were orthographically transcribed by one of the interviewers and cross-checked by an independent native speaker of Slovene. Finally, each utterance was assigned to the participant or interviewer manually by the first author of the present study.
    We note two potential confounding factors. First, the patient recordings were delivered in a different file format (m4a) than the control group (WAV), which may have introduced confounds during pre-processing (e.g., silence detection may behave differently across formats). Second, the experimenter differed between the two groups, which may affect language use due to conversational alignment effects \citep{pickering2004toward, freud2018speech}. We return to these issues in Section~\ref{sec:limitations}.

    \item[Korean:] The Korean dataset was obtained from the Kang corpus, available on DementiaBank \cite{macwhinney2011}. The Kang corpus includes participants with MCI and HCs. Each participant took part in a structured interview consisting of 16 questions. The corpus includes manual transcriptions.
\end{description}

\begin{table*}[t]
\centering
\small
\begin{tabular}{lccccc}
\toprule
Dataset & Condition & Patients & Controls & Total & Patient/Control (\%) \\
\midrule
English  & AD  & 78 & 78 & 156 & 50.0 / 50.0 \\
Slovene  & AD  & 12 & 15 & 27  & 44.4 / 55.6 \\
Korean   & MCI & 40 & 37 & 77  & 51.9 / 48.1 \\
\midrule
\textbf{Total} & -- & \textbf{130} & \textbf{130} & \textbf{260} & \textbf{50.0 / 50.0} \\
\bottomrule
\end{tabular}
\caption{Participant statistics per dataset with within-dataset class proportions. AD: Alzheimer’s Disease; MCI: Mild Cognitive Impairment.}
\label{tab:participants}
\end{table*}

\subsection{Transcript pre-processing}
\label{pre-processing}

For each dataset, we extracted the participant utterances and removed non-orthographic annotations (e.g., the use of `(.)' to indicate short pauses in the Pitt corpus).

For the English and Slovene datasets, each utterance was processed using that language's model from the Stanza NLP library \cite{qi2020stanza}. After tokenisation, tokens labelled as punctuation were excluded, and for each remaining token we extracted the surface form, lemma, universal part-of-speech (UPOS) tag, dependency relation, and syntactic head index.

For Korean, we used the MeCab morphological analyser for tokenisation and part-of-speech (POS) tagging \cite{Kudo2005MeCabY}. Tokens labelled as punctuation were excluded (SF: sentence-final punctuation; SP: comma/pause; SS: brackets/quotation marks; SE: ellipsis; SO: other symbols), and the remaining POS tags were converted to their UPOS equivalents \cite{park2019new}. Dependency-based features were not extracted for Korean, as MeCab does not provide dependency parsing.

\subsection{Features}
\label{sec:features}

From the preprocessed participant-only transcripts we extracted eleven linguistic features that were reported as indicative of AD-related language change in a recent systematic review \cite{shankar2025}. All features were calculated over all utterances per participant and per task. For Korean, two dependency-based features (idea density and syntactic complexity) could not be computed (see Section~\ref{pre-processing}); these values are treated as missing and handled by the imputation procedure described in Section~\ref{sec:representations}.

\begin{description}
    \item[\textbf{Speech Rate}] The number of words uttered by the participant, divided by the total duration of the task in seconds. In English and Slovene, we divided the number of words by the duration from the start of the first participant utterance to the end of the last participant utterance, without excluding interviewer speech. In Korean, we excluded interviewer speech from the duration because the interviewer took a more active role due to the structured nature of the task. Interviewer speech could not be consistently removed from the total duration in English because accurate time-stamps were not available.

    \item[\textbf{Type-Token Ratio}] The number of unique tokens in the participant's speech divided by the total number of words they uttered.

    \item[\textbf{Repetitiveness}] The mean cosine distance between embeddings (produced by Sentence-BERT; \citealp{reimers2019}) of each consecutive pair of participant utterances.

    \item[\textbf{Coherence}] The mean cosine distance of embeddings (produced by Sentence-BERT; \citealp{reimers2019}) between all possible pairs of different participant utterances.

    \item[\textbf{Familiarity}] The mean familiarity per unique word used by the participant. Familiarity is expressed as the number of occurrences per million words in speech corpora, provided by frequency reference lists for each language \cite{dobrovoljc2018, leech2014, kim2024}.

    \item[\textbf{Idea Density}] The number of main verbs divided by the total number of tokens uttered by the participant. Not computed for Korean (Section~\ref{pre-processing}).

    \item[\textbf{Syntactic Complexity}] The mean maximal syntactic tree depth per participant utterance. Not computed for Korean (Section~\ref{pre-processing}).

    \item[\textbf{Verb Ratio}] The number of verbs divided by the total number of tokens.

    \item[\textbf{Noun Ratio}] The number of nouns divided by the total number of tokens.

    \item[\textbf{Pronoun Ratio}] The number of pronouns divided by the total number of tokens.

    \item[\textbf{Pronoun to Noun Ratio}] The number of pronouns divided by the total number of nouns.
\end{description}

\section{Modelling Methodology}
\label{sec:methodology}

\subsection{Task and Data}

We study binary classification of CI (AD or MCI, depending on the dataset) versus HC from speech-derived inputs.
We evaluate performance separately for three languages: English, Slovene, and Korean (Section~\ref{sec:data}). All experiments are conducted within-language (i.e., training and evaluation never mix languages), and results are reported per language and aggregated across languages.

\subsection{Input Representations}
\label{sec:representations}

We compare three input configurations derived from each sample.

\paragraph{(1) Symbolic linguistic features.}
We use an 11-dimensional vector of expert-assisted textual features:
\[
\mathbf{x}_{\text{feat}} \in \mathbb{R}^{11},
\]
corresponding to the features listed in Section~\ref{sec:features}. For Korean, two feature dimensions are missing and are handled by imputation within each fold (see below).

\paragraph{(2) Embedding-based representation.}
We compute a fixed-dimensional dense embedding from the transcript of the participant's utterances
using a frozen multilingual embedding model (\texttt{google/embedding-gemma-300m}):
\[
\mathbf{x}_{\text{emb}} = f_{\text{emb}}(t), \qquad \mathbf{x}_{\text{emb}} \in \mathbb{R}^{d}.
\]
Embeddings are computed once and reused across all evaluation runs.

\paragraph{(3) Fusion of embeddings and features.}
We consider two fusion strategies that combine the embedding and symbolic
feature modalities.

For \emph{early fusion}, we preprocess each modality independently, re-weight the symbolic features to compensate
for the dimensionality imbalance ($d \gg 11$), and concatenate into a single
vector passed to one classifier:
\[
w = \sqrt{\frac{d}{11}}, \qquad
\mathbf{x}_{\text{early}} = \bigl[\,\tilde{\mathbf{x}}_{\text{emb}}\;;\;
w\cdot\tilde{\mathbf{x}}_{\text{feat}}\,\bigr].
\]

For \emph{late fusion}, we train two independent classifiers of the same
family---one on $\tilde{\mathbf{x}}_{\text{emb}}$ and one on
$\tilde{\mathbf{x}}_{\text{feat}}$---and combine their outputs by averaging the
predicted class probabilities:
\[
\hat{p}_{\text{late}} = \frac{1}{2}\Bigl(
    \hat{p}_{\text{emb}} + \hat{p}_{\text{feat}}
\Bigr), \qquad
\hat{y}_{\text{late}} = \mathbf{1}\bigl[\hat{p}_{\text{late}} \geq 0.5\bigr].
\]
This decision-level combination allows each modality to be preprocessed and
modelled independently before fusion.

\paragraph{Preprocessing.}
To prevent data leakage, all preprocessing steps are fit using training data only within each evaluation fold or episode. We apply median imputation per feature dimension using a \texttt{SimpleImputer} fit on the training split. We standardise each feature dimension using z-score normalisation
(\texttt{StandardScaler}) fit on the training split. For both fusion variants, embeddings and symbolic features are imputed and standardised separately.

\subsection{Models}
\label{sec:models}

We compare tabular foundation models, classical ML baselines, and prompted LLMs.

\subsubsection{Tabular and Foundational Models}

We train the following classifiers per language and representation (embeddings, features, early fusion, and late fusion): TabPFN (foundational in-context tabular classifier), RealMLP (tabular deep learning baseline), Logistic Regression (LR), Random Forest (RF), Support Vector Machines (SVM) with both linear (SVM-Linear) and radial basis function (SVM-RBF) kernels, LightGBM (LGBM), and $k$-nearest neighbours ($k$-NN) with $k \in \{3, 5, 7\}$. For late fusion, each model family is instantiated twice per fold or episode (once per modality) and their predicted probabilities are averaged as described in Section~\ref{sec:representations}.

\subsubsection{LLM-based Classification}

We evaluate two LLMs as direct classifiers:
gpt-oss-20b and med-gemma-27b. Both are used in an inference-only setting via an OpenAI-compatible API endpoint served through vLLM.

We test three prompt variants: (1) transcript-only,
(2) linguistic-only (the 11 symbolic features rendered as a numeric list), and
(3) full-data (the transcript concatenated with the symbolic feature list). Full prompt templates are provided in Appendix~1. 

Models are instructed to output exactly one token: \textsc{Control} or \textsc{Patient}. Where supported by the inference server, we enforce a guided-choice constraint restricting outputs to $\{\textsc{Control},\,\textsc{Patient}\}$. Temperature is set to $0$ for deterministic decoding unless otherwise stated.

\subsection{Evaluation Protocols}
\label{sec:evaluation}

All evaluations are conducted \emph{within-language}. We report per-language performance and aggregated averages across languages. To ensure direct comparability between tabular models and LLMs across all evaluation settings, both model families share the same outer leave-one-out loop and the same episodic sampling scheme for few-shot conditions.

\subsubsection{Full-Data Evaluation (Leave-One-Out)}

For each language $L$, we perform leave-one-out cross-validation over all $n_L$ samples. Each fold holds out exactly one sample $i$ for testing and uses the remaining $n_L - 1$ samples as the training pool. For tabular models, preprocessing is fit on the training pool only and applied to both train and test. For LLMs, no demonstrations are used in the zero-shot condition. We refer to this setting as the \emph{full-data} evaluation. For each language, we also report a majority-class predictor trained and evaluated under the same LOO protocol.

\subsubsection{Few-Shot Episodic Evaluation}

To study few-shot behaviour under a unified and directly comparable protocol, we apply the same outer LOO loop and inner episodic sampling scheme to tabular models.

\begin{table}[H]
\centering
\caption{LOO Macro-F1 across all models and input configurations. Classical ML, tabular foundation models (TFMs), gradient boosting, and LLM zero-shot baselines. Best per language in \textbf{bold}. $\dagger$ = zero-shot (no training data).}
\label{tab:all_loo}
\resizebox{\columnwidth}{!}{%
\setlength{\tabcolsep}{4pt}
\renewcommand{\arraystretch}{1.05}
\begin{tabular}{llccc}
\toprule
\textbf{Method} & \textbf{Input} & \textbf{English} & \textbf{Slovene} & \textbf{Korean} \\
\midrule
\multicolumn{5}{l}{\textit{Non-learning baseline}} \\[2pt]
Majority & --- & 0.333 & 0.357 & 0.342 \\
\midrule
\multicolumn{5}{l}{\textit{LLMs (zero-shot$^\dagger$)}} \\[2pt]
\multirow{3}{*}{MedGemma-27B$^\dagger$} & Full & 0.361 & 0.518 & 0.595 \\
 & Ling. & 0.333 & 0.357 & 0.342 \\
 & Trans. & 0.413 & 0.492 & 0.579 \\
\midrule
\multirow{3}{*}{GPT-OSS-20B$^\dagger$} & Full & 0.413 & 0.617 & 0.609 \\
 & Ling. & 0.500 & 0.555 & 0.349 \\
 & Trans. & 0.556 & 0.603 & 0.621 \\
\midrule
\multicolumn{5}{l}{\textit{Tabular foundation models}} \\[2pt]
\multirow{4}{*}{TabPFN} & Emb & 0.343 & \textbf{0.852} & 0.523 \\
 & Feat & 0.814 & 0.454 & 0.740 \\
 & Early & 0.784 & \textbf{0.852} & 0.792 \\
 & Late & \textbf{0.816} & 0.727 & 0.753 \\
\midrule
\multirow{4}{*}{RealMLP} & Emb & 0.493 & 0.802 & 0.566 \\
 & Feat & 0.746 & 0.682 & 0.692 \\
 & Early & 0.743 & 0.701 & 0.737 \\
 & Late & 0.738 & 0.754 & 0.680 \\
\midrule
\multicolumn{5}{l}{\textit{Gradient boosting}} \\[2pt]
\multirow{4}{*}{LGBM} & Emb & 0.549 & 0.357 & 0.342 \\
 & Feat & 0.782 & 0.357 & 0.621 \\
 & Early & 0.763 & 0.357 & 0.621 \\
 & Late & 0.776 & 0.357 & 0.633 \\
\midrule
\multicolumn{5}{l}{\textit{Classical ML}} \\[2pt]
\multirow{4}{*}{LR} & Emb & 0.549 & \textbf{0.852} & 0.523 \\
 & Feat & 0.807 & 0.540 & 0.739 \\
 & Early & 0.801 & 0.735 & 0.792 \\
 & Late & 0.788 & 0.727 & \textbf{0.805} \\
\midrule
\multirow{4}{*}{RF} & Emb & 0.549 & \textbf{0.852} & 0.523 \\
 & Feat & 0.781 & 0.635 & 0.675 \\
 & Early & 0.665 & \textbf{0.852} & 0.714 \\
 & Late & 0.737 & \textbf{0.852} & 0.714 \\
\midrule
\multirow{4}{*}{SVM (linear)} & Emb & 0.549 & \textbf{0.852} & 0.523 \\
 & Feat & 0.813 & 0.540 & 0.727 \\
 & Early & 0.750 & 0.659 & 0.778 \\
 & Late & 0.769 & 0.814 & 0.789 \\
\midrule
\multirow{4}{*}{SVM (RBF)} & Emb & 0.549 & \textbf{0.852} & 0.523 \\
 & Feat & 0.806 & 0.442 & 0.714 \\
 & Early & 0.738 & \textbf{0.852} & 0.766 \\
 & Late & 0.807 & \textbf{0.852} & 0.712 \\
\midrule
\multirow{4}{*}{$k$NN-3} & Emb & 0.333 & 0.357 & 0.523 \\
 & Feat & 0.712 & 0.508 & 0.726 \\
 & Early & 0.654 & \textbf{0.852} & 0.778 \\
 & Late & 0.698 & 0.250 & 0.476 \\
\midrule
\multirow{4}{*}{$k$NN-5} & Emb & 0.325 & 0.308 & 0.523 \\
 & Feat & 0.750 & 0.463 & 0.674 \\
 & Early & 0.652 & 0.814 & 0.779 \\
 & Late & 0.582 & 0.583 & 0.476 \\
\midrule
\multirow{4}{*}{$k$NN-7} & Emb & 0.410 & \textbf{0.852} & 0.523 \\
 & Feat & 0.762 & 0.365 & 0.673 \\
 & Early & 0.665 & \textbf{0.852} & 0.752 \\
 & Late & 0.543 & 0.735 & 0.500 \\
\bottomrule
\end{tabular}%
}
\end{table}
 For each held-out test sample $i$, we sample $k$ examples per class uniformly at random from the remaining $n_L - 1$ samples (the support set), using only these $2k$ samples for training. We repeat this sampling for
$E = 3$ episodes with different random seeds and aggregate predictions across episodes by majority vote; for models producing calibrated probabilities, we additionally average scores before thresholding. We evaluate $k \in \{1, 2, 3, 5\}$. For tabular models, the $2k$ support samples are used to fit the classifier (with preprocessing fit on the same $2k$ samples). For late fusion, two classifiers are fit on the $2k$ samples (one per modality) and their probabilities are averaged.
\subsubsection{Zero-Shot Evaluation for LLMs}

For each language and prompt modality, we classify each sample independently with no labeled demonstrations. \\
\textbf{Evaluation} For each language, model, and evaluation mode we report Macro-F1. To estimate variability due to episodic sampling and any stochasticity in model training, we repeat the full tabular evaluation pipeline three times with different random seeds. Results are aggregated by Language, Model, Evaluation Mode and reported as mean. LLM evaluations are deterministic at temperature $0$ and are therefore reported as single-run results.

\section{Results and Discussion}
\label{discussion}

Leave-one-out (LOO) Macro-F1 results are reported in Table~\ref{tab:all_loo}. Few-shot results (embeddings-only; $k$ shots per class) are reported in Table~\ref{tab:fewshot}. We discuss the findings by research question.

\begin{table}[H]
\centering
\caption{Few-shot Macro-F1 (embeddings-only, $k$ shots/class, 3 seeds) vs.\ LLM zero-shot reference. Best overall per language in \textbf{bold}.}
\label{tab:fewshot}
\resizebox{\columnwidth}{!}{%
\setlength{\tabcolsep}{5pt}
\renewcommand{\arraystretch}{1.05}
\begin{tabular}{llccc}
\toprule
\textbf{Method} & \boldmath{$k$} & \textbf{English} & \textbf{Slovene} & \textbf{Korean} \\
\midrule
\multicolumn{5}{l}{\textit{LLM zero-shot reference (best across models \& modalities)}} \\[2pt]
Best LLM$^\dagger$ & 0 & \textbf{0.556} & 0.617 & 0.621 \\
\midrule
\multicolumn{5}{l}{\textit{Tabular foundation models}} \\[2pt]
\multirow{4}{*}{TabPFN} & 1 & 0.439 & 0.846 & 0.538 \\
 & 2 & 0.442 & 0.815 & 0.504 \\
 & 3 & 0.436 & \textbf{0.852} & 0.560 \\
 & 5 & 0.472 & \textbf{0.852} & 0.532 \\
\midrule
\multirow{4}{*}{RealMLP} & 1 & 0.404 & \textbf{0.852} & 0.497 \\
 & 2 & 0.461 & 0.815 & 0.522 \\
 & 3 & 0.448 & 0.839 & 0.468 \\
 & 5 & 0.455 & \textbf{0.852} & 0.570 \\
\midrule
\multicolumn{5}{l}{\textit{Classical ML}} \\[2pt]
\multirow{4}{*}{LR} & 1 & 0.400 & \textbf{0.852} & 0.497 \\
 & 2 & 0.431 & \textbf{0.852} & 0.568 \\
 & 3 & 0.444 & \textbf{0.852} & 0.552 \\
 & 5 & 0.449 & \textbf{0.852} & \textbf{0.667} \\
\midrule
\multirow{4}{*}{RF} & 1 & 0.382 & \textbf{0.852} & 0.497 \\
 & 2 & 0.462 & \textbf{0.852} & 0.568 \\
 & 3 & 0.458 & \textbf{0.852} & 0.552 \\
 & 5 & 0.482 & \textbf{0.852} & 0.452 \\
\bottomrule
\end{tabular}%
}
\end{table}

\paragraph{RQ1: LLM zero-shot CI detection.}
Zero-shot LLM performance is generally limited relative to supervised tabular models. The best zero-shot LLM result is GPT-OSS-20B on Korean (0.621 Macro-F1), which is +0.279 above the majority baseline (0.342). On English and Slovene, LLM performance varies substantially by prompt modality; for example, MedGemma-27B collapses to majority-class behaviour in the linguistic-only setting on English (0.333 Macro-F1). Despite being a medical-domain model, MedGemma-27B underperforms GPT-OSS-20B across all three languages, suggesting general instruction-following behaviour and robustness to prompt formatting may matter more than domain specialisation in this setting.

\paragraph{RQ2: Modality sensitivity.}
No single input modality consistently dominates across languages and models. Transcript-only input works best for English and Korean with GPT-OSS-20B, while full-data prompts (transcript + features) do not reliably outperform single-modality prompts. This suggests that, in an inference-only setting, LLMs may struggle to integrate heterogeneous numeric and textual evidence as reliably as simpler tabular fusion approaches.

\paragraph{RQ3: Symbolic features + tabular models vs.\ LLMs.}
Tabular models with symbolic features decisively outperform zero-shot LLM baselines (+0.18 to +0.26 Macro-F1 across languages when comparing best results per language). The 11-feature vector is highly informative: on English, TabPFN achieves 0.814 with features alone, a +0.471 improvement over embeddings-only (0.343). Classical ML baselines remain competitive, with LR achieving the best Korean result (0.805, late fusion). Slovene is an exception where embeddings dominate features (0.852 vs.\ 0.454 for TabPFN), though potential confounds (different recording formats and experimenters across groups) may inflate embedding-based separability (Sections~\ref{sec:data} and \ref{sec:limitations}).

\paragraph{RQ4: Fusion strategies.}
Early fusion generally performs best for Slovene and yields strong performance for Korean, indicating that combining complementary information sources can improve stability across datasets. Features-only is strongest for English, where the engineered linguistic signal is particularly predictive and adding high-dimensional embeddings can dilute that signal for some models. To better understand when fusion helps, we analyse alignment between feature space and embedding space (Table~\ref{tab:alignment}). English and Korean show near-orthogonal representations (CKA 0.024 and 0.016; low Overlap@5), consistent with fusion providing complementary information. Slovene exhibits higher alignment (CKA 0.181; Overlap@5 0.200), suggesting that in this dataset embeddings may already encode much of the feature-level signal (or reflect dataset-specific confounds).

\paragraph{Few-shot vs.\ zero-shot LLMs.}
Even with minimal labels, tabular models can match or exceed LLM zero-shot performance for some languages. For Slovene, embedding-based few-shot models exceed the best LLM from just $k$=1 example per class (0.846--0.852 vs.\ 0.617). For Korean, LR reaches 0.667 at $k$=5, exceeding the best LLM (0.621). In English, the best LLM zero-shot result (0.556) remains higher than embedding-only few-shot baselines, highlighting the importance of incorporating symbolic features and/or stronger representations for small-data supervision.

\begin{table}[H]
\centering
\caption{Feature--embedding space alignment. Low CKA and Spearman $\rho$ indicate weak alignment between representations; Procrustes disparity~\cite{williams2021generalized} near 2.0 indicates geometric dissimilarity. Overlap@5 = shared $k$NN neighbours; Purity@5 = same-class neighbours.}
\label{tab:alignment}
\resizebox{\columnwidth}{!}{%
\setlength{\tabcolsep}{4pt}
\renewcommand{\arraystretch}{1.05}
\begin{tabular}{lcccccc}
\toprule
\textbf{Language} & \textbf{CKA} & \textbf{Spearman $\rho$} & \textbf{Procrustes} & \textbf{Overlap@5} & \textbf{Purity\textsubscript{feat}@5} & \textbf{Purity\textsubscript{emb}@5} \\
\midrule
English  & 0.024 & 0.001  & 1.758 & 0.032 & 0.658 & 0.501 \\
Slovene  & 0.181 & 0.116  & 1.432 & 0.200 & 0.511 & 0.563 \\
Korean   & 0.016 & 0.005  & 1.835 & 0.049 & 0.610 & 0.610 \\
\bottomrule
\end{tabular}%
}
\end{table}

\section{Conclusion}

We evaluated CI detection across three languages (English, Slovene, and Korean) comparing LLM zero-shot prompting, tabular foundation models, and classical ML. LLMs provide usable no-training baselines (best Macro-F1: 0.621), but supervised tabular models—especially those using expert-assisted symbolic features and fusion—achieve substantially higher performance (+0.18 to +0.26 Macro-F1 over the best LLM per language). Across datasets, lightweight classical models remain highly competitive, with TabPFN reaching 0.816 on English (late fusion) and LR reaching 0.805 on Korean (late fusion). Alignment analysis indicates that feature and embedding representations are weakly aligned in English and Korean, providing principled support for fusion; Slovene shows higher alignment, consistent with embeddings dominating in that dataset. For practical deployment in small-data CI detection, tabular models operating on transparent linguistic markers offer a strong and interpretable alternative to inference-only LLM classification, while LLM prompting remains a useful reference point when training data are unavailable.

\section{Code Availability}
The source code is publicly available at \url{https://github.com/bkolosk1/foundational-ci-detection}.

\section{Limitations}
\label{sec:limitations}

The Slovene dataset ($n$=27) exhibits potential confounds: different recording formats and experimenters for patient and control groups may inflate embedding-based performance. Two symbolic features (idea density and syntactic complexity) are unavailable for Korean due to parser limitations, and are therefore treated as missing and imputed; this reduces the amount of available symbolic information for that language. While the symbolic features are individually interpretable, fusion with 768-dimensional embeddings reduces transparency; future work should investigate explanation methods (e.g., SHAP) for fusion models and assess robustness across elicitation paradigms. The relatively small dataset size, even if it is typical for this field, restricts the strength and generalisability of our conclusions. Moreover, interviewer speech was not excluded from the speech-rate calculation in the English and Slovene data, which may have affected these measures. As a result, the reported speech-rate values may not fully reflect participant speech alone and should be interpreted with caution. A further limitation is that the study does not examine potential sources of bias related to participant characteristics such as age, education, and linguistic background, all of which may affect linguistic performance and, in turn, influence classification outcomes. It also does not account for differences in cognitive or brain reserve. Assessing such demographic and individual differences is essential before any clinical deployment. Finally, two of the three datasets use picture description tasks; generalisation to other elicitation paradigms (spontaneous speech, narrative recall) remains to be investigated.

\section*{Acknowledgments}

We acknowledge the financial support from the Slovenian Research Agency ARIS via the projects Cross-Lingual Analysis for Detection of Cognitive Impairment in Less-Resourced Languages (CroDeCo; J6-60109), and Natural Language Processing for Corpus Analysis in the Medical Humanities (BI-VB/25-27-021)
, and research core funding for the programme Knowledge Technologies (P2-0103).

BK is funded by the Young Researcher Grant PR-12394, and JC by the Young Researcher Grant PR-13409.

The English dataset was based on the Pitt Corpus \cite{becker1994}, which was produced with the support of grants NIA AG03705 and AG05133 to the original authors of the corpus.

\section{Bibliographical References}\label{sec:reference}

\bibliographystyle{lrec2026-natbib}
\bibliography{lrec2026-example}

@article{ray2014dementia,
  author  = {Ray, Sujata and Davidson, Susan},
  title   = {Dementia and cognitive decline. A review of the evidence},
  journal = {Age UK},
  volume  = {27},
  pages   = {10--12},
  year    = {2014}
}

@article{petersen2016mild,
  author  = {Petersen, Ronald C.},
  title   = {Mild cognitive impairment},
  journal = {CONTINUUM: Lifelong Learning in Neurology},
  volume  = {22},
  number  = {2},
  pages   = {404--418},
  year    = {2016},
  publisher={LWW}
}

@article{morley2018,
  author  = {Morley, John E},
  title   = {An overview of cognitive impairment},
  journal = {Clinics in Geriatric Medicine},
  volume  = {34},
  number  = {4},
  pages   = {505--513},
  year    = {2018}
}

@article{livingston2024,
  author  = {Livingston, Gill and Huntley, Jonathan and Liu, Kathy Y and Costafreda, Sergi G and Selb{\ae}k, Geir and Alladi, Suvarna and Ames, David and Banerjee, Sube and Burns, Alistair and Brayne, Carol and others},
  title   = {The Lancet Commissions},
  journal = {Lancet},
  volume  = {404},
  pages   = {572--628},
  year    = {2024}
}

@article{nasreddine2005,
  author  = {Nasreddine, Ziad S and Phillips, Natalie A and B{\'e}dirian, Val{\'e}rie and Charbonneau, Simon and Whitehead, Victor and Collin, Isabelle and Cummings, Jeffrey L and Chertkow, Howard},
  title   = {The Montreal Cognitive Assessment, MoCA: a brief screening tool for mild cognitive impairment},
  journal = {Journal of the American Geriatrics Society},
  volume  = {53},
  number  = {4},
  pages   = {695--699},
  year    = {2005}
}

@article{obryant2008,
  author  = {O’Bryant, Sid E and Waring, Stephen C and Cullum, C Munro and Hall, James and Lacritz, Laura and Massman, Paul J and Lupo, Philip J and Reisch, Joan S and Doody, Rachelle and Texas Alzheimer's Research Consortium and others},
  title   = {Staging dementia using Clinical Dementia Rating Scale Sum of Boxes scores: a Texas Alzheimer's research consortium study},
  journal = {Archives of Neurology},
  volume  = {65},
  number  = {8},
  pages   = {1091--1095},
  year    = {2008}
}

@article{hampel2018,
  author  = {Hampel, Harald and O’Bryant, Sid E and Molinuevo, Jos{\'e} L and Zetterberg, Henrik and Masters, Colin L and Lista, Simone and Kiddle, Steven J and Batrla, Richard and Blennow, Kaj},
  title   = {Blood-based biomarkers for Alzheimer disease: mapping the road to the clinic},
  journal = {Nature Reviews Neurology},
  volume  = {14},
  number  = {11},
  pages   = {639--652},
  year    = {2018}
}

@article{mckhann2011,
  author  = {McKhann, Guy M and Knopman, David S and Chertkow, Howard and Hyman, Bradley T and Jack Jr, Clifford R and Kawas, Claudia H and Klunk, William E and Koroshetz, Walter J and Manly, Jennifer J and Mayeux, Richard and others},
  title   = {The diagnosis of dementia due to Alzheimer's disease: recommendations from the National Institute on Aging-Alzheimer's Association workgroups on diagnostic guidelines for Alzheimer's disease},
  journal = {Alzheimer's \& Dementia},
  volume  = {7},
  number  = {3},
  pages   = {263--269},
  year    = {2011}
}

@article{taler2008,
  author  = {Taler, Vanessa and Phillips, Natalie A},
  title   = {Language performance in Alzheimer's disease and mild cognitive impairment: a comparative review},
  journal = {Journal of Clinical and Experimental Neuropsychology},
  volume  = {30},
  number  = {5},
  pages   = {501--556},
  year    = {2008}
}

@article{fraser2015,
  author  = {Fraser, Kathleen C and Meltzer, Jed A and Rudzicz, Frank},
  title   = {Linguistic features identify Alzheimer’s disease in narrative speech},
  journal = {Journal of Alzheimer’s Disease},
  volume  = {49},
  number  = {2},
  pages   = {407--422},
  year    = {2015}
}

@article{boschi2017,
  author  = {Boschi, Veronica and Catricala, Eleonora and Consonni, Monica and Chesi, Cristiano and Moro, Andrea and Cappa, Stefano F},
  title   = {Connected speech in neurodegenerative language disorders: a review},
  journal = {Frontiers in Psychology},
  volume  = {8},
  pages   = {269},
  year    = {2017}
}

@book{goodglass1983,
  author  = {Goodglass, Harold and Kaplan, Edith},
  title   = {The assessment of aphasia and related disorders},
  year    = {1983}
}

@article{trzepacz2015,
  author  = {Trzepacz, Paula T and Hochstetler, Helen and Wang, Shufang and Walker, Brett and Saykin, Andrew J and Alzheimer’s Disease Neuroimaging Initiative},
  title   = {Relationship between the Montreal Cognitive Assessment and Mini-mental State Examination for assessment of mild cognitive impairment in older adults},
  journal = {BMC Geriatrics},
  volume  = {15},
  number  = {1},
  pages   = {107},
  year    = {2015}
}

@article{slegers2018,
  author  = {Slegers, Antoine and Filiou, Renee-Pier and Montembeault, Maxime and Brambati, Simona Maria},
  title   = {Connected speech features from picture description in Alzheimer’s disease: A systematic review},
  journal = {Journal of Alzheimer’s Disease},
  volume  = {65},
  number  = {2},
  pages   = {519--542},
  year    = {2018}
}

@article{ramoshenderson2025,
  author  = {Ramos-Henderson, Miguel and Calder{\'o}n, Carlos and Domic-Siede, Marcos},
  title   = {Education bias in typical brief cognitive tests used for the detection of dementia in elderly population with low educational level: a critical review},
  journal = {Applied Neuropsychology: Adult},
  volume  = {32},
  number  = {1},
  pages   = {253--261},
  year    = {2025}
}

@article{patnode2020,
  author  = {Patnode, Carrie D and Perdue, Leslie A and Rossom, Rebecca C and Rushkin, Megan C and Redmond, Nadia and Thomas, Rachel G and Lin, Jennifer S},
  title   = {Screening for cognitive impairment in older adults: updated evidence report and systematic review for the US Preventive Services Task Force},
  journal = {JAMA},
  volume  = {323},
  number  = {8},
  pages   = {764--785},
  year    = {2020}
}

@article{delafuentegarcia2020,
  author  = {De la Fuente Garcia, Sofia and Ritchie, Craig W and Luz, Saturnino},
  title   = {Artificial intelligence, speech, and language processing approaches to monitoring Alzheimer’s disease: a systematic review},
  journal = {Journal of Alzheimer’s Disease},
  volume  = {78},
  number  = {4},
  pages   = {1547--1574},
  year    = {2020}
}

@inproceedings{pahar2025,
  author    = {Pahar, Madhurananda and Tao, Fuxiang and Mirheidari, Bahman and Pevy, Nathan and Bright, Rebecca and Gadgil, Swapnil and Sproson, Lise and Braun, Dorota and Illingworth, Caitlin and Blackburn, Daniel and others},
  title     = {CognoSpeak: an automatic, remote assessment of early cognitive decline in real-world conversational speech},
  booktitle = {2025 IEEE Symposium on Computational Intelligence in Health and Medicine (CIHM)},
  pages     = {1--7},
  year      = {2025},
  publisher = {IEEE}
}

@article{guan2025,
  author  = {Guan, Hao and Novoa-Laurentiev, John and Zhou, Li},
  title   = {CD-Tron: Leveraging large clinical language model for early detection of cognitive decline from electronic health records},
  journal = {Journal of Biomedical Informatics},
  pages   = {104830},
  year    = {2025}
}

@article{mao2022,
  author  = {Mao, Chengsheng and Xu, Jie and Rasmussen, Luke and Li, Yikuan and Adekkanattu, Prakash and Pacheco, Jennifer and Bonakdarpour, Borna and Vassar, Robert and Jiang, Guoqian and Wang, Fei and others},
  title   = {AD-BERT: using pre-trained contextualized embeddings to predict the progression from mild cognitive impairment to Alzheimer's disease},
  journal = {arXiv preprint arXiv:2212.06042},
  year    = {2022}
}

@article{huang2024,
  author  = {Huang, Lihe and Yang, Hao and Che, Yiran and Yang, Jingjing},
  title   = {Automatic speech analysis for detecting cognitive decline of older adults},
  journal = {Frontiers in Public Health},
  volume  = {12},
  pages   = {1417966},
  year    = {2024}
}

@article{luz2021,
  author  = {Luz, Saturnino and Haider, Fasih and de la Fuente Garcia, Sofia and Fromm, Davida and MacWhinney, Brian},
  title   = {Alzheimer's dementia recognition through spontaneous speech},
  journal = {Frontiers in Computer Science},
  volume  = {3},
  pages   = {780169},
  year    = {2021}
}

@article{kaser2024,
  author  = {Kaser, Alyssa N and Lacritz, Laura H and Winiarski, Holly R and Gabirondo, Peru and Schaffert, Jeff and Coca, Alberto J and Jim{\'e}nez-Raboso, Javier and Rojo, Tomas and Zaldua, Carla and Honorato, Iker and others},
  title   = {A novel speech analysis algorithm to detect cognitive impairment in a Spanish population},
  journal = {Frontiers in Neurology},
  volume  = {15},
  pages   = {1342907},
  year    = {2024}
}

@article{ilias2022,
  author  = {Ilias, Loukas and Askounis, Dimitris},
  title   = {Explainable identification of dementia from transcripts using transformer networks},
  journal = {IEEE Journal of Biomedical and Health Informatics},
  volume  = {26},
  number  = {8},
  pages   = {4153--4164},
  year    = {2022}
}

@inproceedings{zhu2022,
  author    = {Zhu, Youxiang and Liang, Xiaohui and Batsis, John A and Roth, Robert M},
  title     = {Domain-aware intermediate pretraining for dementia detection with limited data},
  booktitle = {Interspeech},
  volume    = {2022},
  pages     = {2183},
  year      = {2022}
}

@article{dearribaperez2024,
  author  = {de Arriba-P{\'e}rez, Francisco and Garc{\'\i}a-M{\'e}ndez, Silvia and Otero-Mosquera, Javier and Gonz{\'a}lez-Casta{\~n}o, Francisco J},
  title   = {Explainable cognitive decline detection in free dialogues with a Machine Learning approach based on pre-trained Large Language Models},
  journal = {arXiv preprint arXiv:2411.02036},
  year    = {2024}
}

@inproceedings{zheng2024,
  author    = {Zheng, Tian and Xie, Xurong and Peng, Xiaolan and Chen, Hui and Tian, Feng},
  title     = {Alzheimer’s Disease Detection Based on Large Language Model Prompt Engineering},
  booktitle = {International Conference on Social Robotics},
  pages     = {207--216},
  year      = {2024},
  publisher = {Springer Nature Singapore}
}

@inproceedings{botelho2024,
  author    = {Botelho, Catarina and Mendon{\c{c}}a, John and Pompili, Anna and Schultz, Tanja and Abad, Alberto and Trancoso, Isabel},
  title     = {Macro-descriptors for Alzheimer's disease detection using large language models},
  booktitle = {Interspeech},
  pages     = {1975--1979},
  year      = {2024},
  doi       = {10.21437/Interspeech.2024-1255}
}

@article{berube2019,
  author  = {Berube, Shauna and Nonnemacher, Jodi and Demsky, Cornelia and Glenn, Shenly and Saxena, Sadhvi and Wright, Amy and Tippett, Donna C and Hillis, Argye E},
  title   = {Stealing cookies in the twenty-first century: Measures of spoken narrative in healthy versus speakers with aphasia},
  journal = {American Journal of Speech-Language Pathology},
  volume  = {28},
  number  = {1S},
  pages   = {321--329},
  year    = {2019}
}

@misc{boersma2021,
  author       = {Boersma, Paul and Weenink, David},
  title        = {Praat: Doing phonetics by computer (Version 6.4.06) [Computer software]},
  year         = {2021},
  howpublished = {\url{http://www.praat.org/}}
}

@article{shankar2025,
  author  = {Shankar, Ravi and Bundele, Anjali and Mukhopadhyay, Amartya},
  title   = {A Systematic Review of Natural Language Processing Techniques for Early Detection of Cognitive Impairment},
  journal = {Mayo Clinic Proceedings: Digital Health},
  volume  = {3},
  number  = {2},
  pages   = {100205},
  year    = {2025},
  doi     = {10.1016/j.mcpdig.2025.100205}
}

@inproceedings{reimers2019,
  author    = {Reimers, Nils and Gurevych, Iryna},
  title     = {Sentence-BERT: Sentence Embeddings using Siamese BERT-Networks},
  booktitle = {Proceedings of the 2019 Conference on Empirical Methods in Natural Language Processing},
  publisher = {Association for Computational Linguistics},
  year      = {2019}
}

@misc{dobrovoljc2018,
  author       = {Dobrovoljc, Kaja},
  title        = {Gos corpus n-grams 2.0},
  year         = {2018},
  howpublished = {\url{https://www.clarin.si/repository/xmlui/handle/11356/1195}}
}

@book{leech2014,
  author    = {Leech, Geoffrey and Rayson, Paul and others},
  title     = {Word Frequencies in Written and Spoken English: Based on the British National Corpus},
  publisher = {Routledge},
  year      = {2014},
  url       = {https://doi.org/10.4324/9781315840161}
}

@inproceedings{kim2024,
  author    = {Kim, Jin-seo and Choi, Anna Seo Gyeong and Cho, Sunghye},
  title     = {KoFREN: Comprehensive Korean Word Frequency Norms Derived from Large Scale Free Speech Corpora},
  booktitle = {Proceedings of the Joint Conference on Language Resources and Evaluation},
  year      = {2024}
}

@article{jiang2026llms,
  title={What Do LLMs Know About Alzheimer's Disease? Fine-Tuning, Probing, and Data Synthesis for AD Detection},
  author={Jiang, Lei and Zhou, Yue and Parde, Natalie},
  journal={arXiv preprint arXiv:2602.11177},
  year={2026}
}

@article{macwhinney2011,
  author  = {MacWhinney, Brian and Fromm, Davida and Forbes, Margaret and Holland, Audrey},
  title   = {AphasiaBank: Methods for studying discourse},
  journal = {Aphasiology},
  volume  = {25},
  pages   = {1286--1307},
  year    = {2011}
}

@inproceedings{luz2020address,
  author    = {Luz, Saturnino and Haider, Farhana and de la Fuente, Sofia and Fromm, Davida and MacWhinney, Brian},
  title     = {Alzheimer's Dementia Recognition through Spontaneous Speech: The ADReSS Challenge},
  booktitle = {Proceedings of Interspeech 2020},
  pages     = {2172--2176},
  year      = {2020}
}

@article{becker1994,
  author  = {Becker, James T. and Boller, Francois and Lopez, Oscar L. and Saxton, Juliana and McGonigle, Kenneth L.},
  title   = {The natural history of Alzheimer's disease: Description of study cohort and accuracy of diagnosis},
  journal = {Archives of Neurology},
  volume  = {51},
  number  = {6},
  pages   = {585--594},
  year    = {1994}
}

@misc{ding2026longitudinalprogressionpredictionalzheimers,
      title={Longitudinal Progression Prediction of Alzheimer's Disease with Tabular Foundation Model}, 
      author={Yilang Ding and Jiawen Ren and Jiaying Lu and Gloria Hyunjung Kwak and Armin Iraji and Shengpu Tang and Alex Fedorov},
      year={2026},
      eprint={2508.17649},
      archivePrefix={arXiv},
      primaryClass={cs.LG},
      url={https://arxiv.org/abs/2508.17649}, 
}

@article{hollmann2025accurate,
  title={Accurate predictions on small data with a tabular foundation model},
  author={Hollmann, Noah and M{\"u}ller, Samuel and Purucker, Lennart and Krishnakumar, Arjun and K{\"o}rfer, Max and Hoo, Shi Bin and Schirrmeister, Robin Tibor and Hutter, Frank},
  journal={Nature},
  volume={637},
  number={8045},
  pages={319--326},
  year={2025},
  publisher={Nature Publishing Group UK London}
}

@article{tran2024predicting,
  title={Predicting dementia in Parkinson's disease on a small tabular dataset using hybrid LightGBM--TabPFN and SHAP},
  author={Tran, Vinh Quang and Byeon, Haewon},
  journal={Digital Health},
  volume={10},
  pages={20552076241272585},
  year={2024},
  publisher={SAGE Publications Sage UK: London, England}
}

@article{pickering2004toward,
  title={Toward a mechanistic psychology of dialogue},
  author={Pickering, Martin J and Garrod, Simon},
  journal={Behavioral and brain sciences},
  volume={27},
  number={2},
  pages={169--190},
  year={2004},
  publisher={Cambridge University Press}
}

@inproceedings{qi2020stanza,
    title={Stanza: A {Python} Natural Language Processing Toolkit for Many Human Languages},
    author={Qi, Peng and Zhang, Yuhao and Zhang, Yuhui and Bolton, Jason and Manning, Christopher D.},
    booktitle = "Proceedings of the 58th Annual Meeting of the Association for Computational Linguistics: System Demonstrations",
    year={2020}
}

@inproceedings{Kudo2005MeCabY,
  title={MeCab : Yet Another Part-of-Speech and Morphological Analyzer},
  author={Takumitsu Kudo},
  year={2005},
  url={https://api.semanticscholar.org/CorpusID:61584143}
}

@article{freud2018speech,
  title={Speech rate adjustment of adults during conversation},
  author={Freud, Debora and Ezrati-Vinacour, Ruth and Amir, Ofer},
  journal={Journal of fluency disorders},
  volume={57},
  pages={1--10},
  year={2018},
  publisher={Elsevier}
}

@inproceedings{williams2021generalized,
  title={Generalized shape metrics on neural representations},
  author={Williams, Alex H and Kunz, Erin and Kornblith, Simon and Linderman, Scott W},
  booktitle={Advances in Neural Information Processing Systems},
  volume={34},
  pages={4738--4750},
  year={2021}
}

@inproceedings{park2019new,
  title={A new annotation scheme for the Sejong part-of-speech tagged corpus},
  author={Park, Jungyeul and Tyers, Francis},
  booktitle={Proceedings of the 13th Linguistic Annotation Workshop},
  pages={195--202},
  year={2019}
}

\bibliographystylelanguageresource{lrec2026-natbib}
\bibliographylanguageresource{languageresource}

\section*{Appendix 1: Prompt templates}\label{sec:prompts}

All LLM-based classifications use the same system instruction and output
constraint, differing only in the data fields provided to the model. The system
prompt and output format are shared across all three variants:

\begin{tcolorbox}[
    colback=gray!6,
    colframe=gray!40,
    title={\small\textbf{System Prompt (all variants)}},
    fonttitle=\bfseries,
    boxrule=0.4pt,
    left=6pt, right=6pt, top=4pt, bottom=4pt
]
\small
You are a binary classifier for a research dataset (non-diagnostic).
Use only the provided transcript and/or linguistic metrics.
Inputs may be English, Slovene, or Korean; treat
multilingualism, accent, dialect, and topical content as neutral.
Ignore demographic/identity attributes and stereotypes.
Assume no class base-rate. Do not reveal reasoning.\\[4pt]
\textbf{Output:} Exactly one word --- \texttt{Control} or \texttt{Patient}.
\end{tcolorbox}

\vspace{0.5em}
\noindent
The three prompt variants differ in which data fields are included in the
query block:

\begin{tcolorbox}[
    colback=blue!4,
    colframe=blue!30,
    title={\small\textbf{Variant 1 — Transcript-only}},
    fonttitle=\bfseries,
    boxrule=0.4pt,
    left=6pt, right=6pt, top=4pt, bottom=4pt
]
\small
\texttt{[DATA INPUT FOR ID: \{id\} \quad Language: \{language\}]}\\[4pt]
\texttt{[TRANSCRIPT]}\\
\textit{\{transcript\_patient\}}\\[6pt]
\texttt{[INSTRUCTIONS]}\\
Classify strictly as ``Control'' or ``Patient'' using only evidence
present in the provided fields. Output exactly one word.\\[6pt]
\texttt{[OUTPUT FORMAT]}\\
Return exactly one word: Control or Patient.
\end{tcolorbox}

\vspace{0.5em}

\begin{tcolorbox}[
    colback=green!4,
    colframe=green!35,
    title={\small\textbf{Variant 2 — Linguistic-only}},
    fonttitle=\bfseries,
    boxrule=0.4pt,
    left=6pt, right=6pt, top=4pt, bottom=4pt
]
\small
\texttt{[DATA INPUT FOR ID: \{id\} \quad Language: \{language\}]}\\[4pt]
\texttt{[LINGUISTIC METRICS]}\\
\begin{tabular}{ll}
- Speech Rate:            & \textit{\{value\}} \\
- Ttr:                    & \textit{\{value\}} \\
- Noun Ratio:             & \textit{\{value\}} \\
- Verb Ratio:             & \textit{\{value\}} \\
- Pronoun Ratio:          & \textit{\{value\}} \\
- Pronoun To Noun Ratio:  & \textit{\{value\}} \\
- Mean Frequency:         & \textit{\{value\}} \\
- Coherence:              & \textit{\{value\}} \\
- Repetitiveness:         & \textit{\{value\}} \\
- Idea Density:           & \textit{\{value\}} \\
- Syntactic Complexity:   & \textit{\{value\}} \\
\end{tabular}\\[6pt]
\texttt{[INSTRUCTIONS]}\\
Classify strictly as ``Control'' or ``Patient'' using only evidence
present in the provided fields. Output exactly one word.\\[6pt]
\texttt{[OUTPUT FORMAT]}\\
Return exactly one word: Control or Patient.
\end{tcolorbox}

\vspace{0.5em}

\begin{tcolorbox}[
    colback=orange!4,
    colframe=orange!35,
    title={\small\textbf{Variant 3 — Full-data (Transcript + Linguistic Metrics)}},
    fonttitle=\bfseries,
    boxrule=0.4pt,
    left=6pt, right=6pt, top=4pt, bottom=4pt
]
\small
\texttt{[DATA INPUT FOR ID: \{id\} \quad Language: \{language\}]}\\[4pt]
\texttt{[TRANSCRIPT]}\\
\textit{\{transcript\_patient\}}\\[6pt]
\texttt{[LINGUISTIC METRICS]}\\
\begin{tabular}{ll}
- Speech Rate:            & \textit{\{value\}} \\
- Ttr:                    & \textit{\{value\}} \\
- Noun Ratio:             & \textit{\{value\}} \\
- Verb Ratio:             & \textit{\{value\}} \\
- Pronoun Ratio:          & \textit{\{value\}} \\
- Pronoun To Noun Ratio:  & \textit{\{value\}} \\
- Mean Frequency:         & \textit{\{value\}} \\
- Coherence:              & \textit{\{value\}} \\
- Repetitiveness:         & \textit{\{value\}} \\
- Idea Density:           & \textit{\{value\}} \\
- Syntactic Complexity:   & \textit{\{value\}} \\
\end{tabular}\\[6pt]
\texttt{[INSTRUCTIONS]}\\
Classify strictly as ``Control'' or ``Patient'' using only evidence
present in the provided fields. Output exactly one word.\\[6pt]
\texttt{[OUTPUT FORMAT]}\\
Return exactly one word: Control or Patient.
\end{tcolorbox}

\vspace{0.5em}
\noindent
For few-shot variants, a \texttt{[EXAMPLES (labeled)]} block is prepended
before the query, containing $2k$ demonstration cases (one per support
sample), each formatted identically to the query block above and annotated
with their ground-truth label (\texttt{Label: Control} or
\texttt{Label: Patient}).

\end{document}